\documentclass[10pt]{article}
\usepackage{makeidx}
\usepackage{multirow}
\usepackage{multicol}
\usepackage[multi-part-units=single]{siunitx}
\usepackage[dvipsnames,svgnames,table]{xcolor}
\usepackage{graphicx}
\usepackage{epstopdf}
\usepackage{ulem}
\usepackage{hyperref}
\usepackage{amsmath}
\usepackage{amssymb}
\usepackage{bm}

\author{DrX}
\title{Towards Deep Learning-Based Electrode Tracking Using Automatically Generated Weak Labels}
\usepackage[paperwidth=595pt,paperheight=842pt,top=70pt,right=56pt,bottom=70pt,left=56pt]{geometry}

\makeatletter
	{\par\setlength{\parindent}{#3}
	\setlength{\leftmargin}{#1}       \setlength{\rightmargin}{#2}%
	\advance\linewidth -\leftmargin       \advance\linewidth -\rightmargin%
	\advance\@totalleftmargin\leftmargin  \@setpar{{\@@par}}%
	\parshape 1\@totalleftmargin \linewidth\ignorespaces}{\par}%
\makeatother 

\begin{document}

\pagestyle{empty}

\begin{center}
\textbf{\Large{Towards Deep Learning-Based EEG Electrode Detection Using Automatically Generated Labels}}
\end{center}

\begin{center}
\textit{N. Gessert${^1}$, M. Gromniak${^1}$, M. Bengs${^1}$, L. Matth\"aus${^2}$, A. Schlaefer${^1}$}
\end{center}

\begin{center}
\textit{${^1}$ Institute of Medical Technology, Hamburg University of Technology, Hamburg, Germany}\\
\textit{${^2}$ eemagine Medical Imaging Solutions GmbH, Berlin, Germany}\\
\end{center}

\begin{center}
\text{Contact: nils.gessert@tuhh.de}
\end{center}

{\raggedright
\textbf{\textit{Abstract}}
}
\\
\\ 
\textit{Electroencephalography (EEG) allows for source measurement of electrical brain activity. Particularly for inverse localization, the electrode positions on the scalp need to be known. Often, systems such as optical digitizing scanners are used for accurate localization with a stylus. However, the approach is time-consuming as each electrode needs to be scanned manually and the scanning systems are expensive. We propose using an RGBD camera to directly track electrodes in the images using deep learning methods. Studying and evaluating deep learning methods requires large amounts of labeled data. To overcome the time-consuming data annotation, we generate a large number of ground-truth labels using a robotic setup. We demonstrate that deep learning-based electrode detection is feasible with a mean absolute error of \SI[separate-uncertainty = true]{5.69 \pm 6.1}{\milli\metre} and that our annotation scheme provides a useful environment for studying deep learning methods for electrode detection.}

\section{\hspace{14pt}Problem}

Electroencephalography (EEG) is a method that allows for measuring electrical brain activity, e.g., to assess patients' motor function impairment or monitor progress in patients' recovery process \cite{otten2015framework}. For accurate brain current estimation based on the measured signals on the scalp, knowledge of the electrodes' location is required \cite{plummer2008eeg}. 

A typical method for electrode placement is the 10-20 system \cite{jasper1958report} or its refined variants \cite{jurcak200710} where the positions are determined based on anatomical landmarks. Identification of anatomical landmarks relies on visual inspection and palpation by the practitioner which is error-prone. Instead, using accurate localization systems, e.g., using optical digitizing scanners with a stylus \cite{towle1993spatial} or MRI-based localization have been proposed \cite{brinkmann1998scalp}. Often, these systems are expensive and recording all electrodes' location is time-consuming. Therefore, photogrammetric methods have been proposed where a single \cite{qian2011single} or multiple cameras \cite{reis2015using} are used to localize the electrodes on the head. These methods are advantageous as cheap cameras can be used for accurate localization. Recent methods often rely on depth (time-of-flight) and/or multiple RGB images for reconstruction of the 3D electrode positions \cite{clausner2017photogrammetry}. 

Previous photogrammetric methods come with two major drawbacks. First, computer vision techniques for 3D reconstruction and electrode detection rely on handcrafted features and algorithms which are often limited to the specific scenarios they were engineered for and the algorithms often come with long execution times. Second, previous approaches usually assume a fixed head location. Both hinder application in mobile and changing environments such as ambulances where head movement is inevitable and fast detection is needed. Thus, fast algorithms that deal with large head pose variation are required. In recent years, deep learning methods have shown remarkable performance for a variety of computer vision tasks such as real-time object detection \cite{ren2015faster} and head pose estimation \cite{ranjan2019hyperface}. In this paper we study the feasibility of electrode detection using convolutional neural networks (CNNs). To facilitate and study this approach, large amounts of annotated data are required. Therefore, we propose a setup using a robot with a head phantom attached to the robot's endeffector. An EEG electrode cap is placed on the head phantom. Then, an RGBD camera acquires images of the head phantom which is moved to different positions and orientations. The electrodes are first labeled in a single image. Then, the electrode locations are transformed to each head pose using a hand-eye calibration and the initial markings. We study whether these automatically generated labels can be learned by a CNN which directly predicts the electrode locations from the images.
\section{\hspace{14pt}Material and Methods}

\subsection{Experimental Setup}

\begin{figure}
	\centering
	\includegraphics [width=0.8\linewidth]{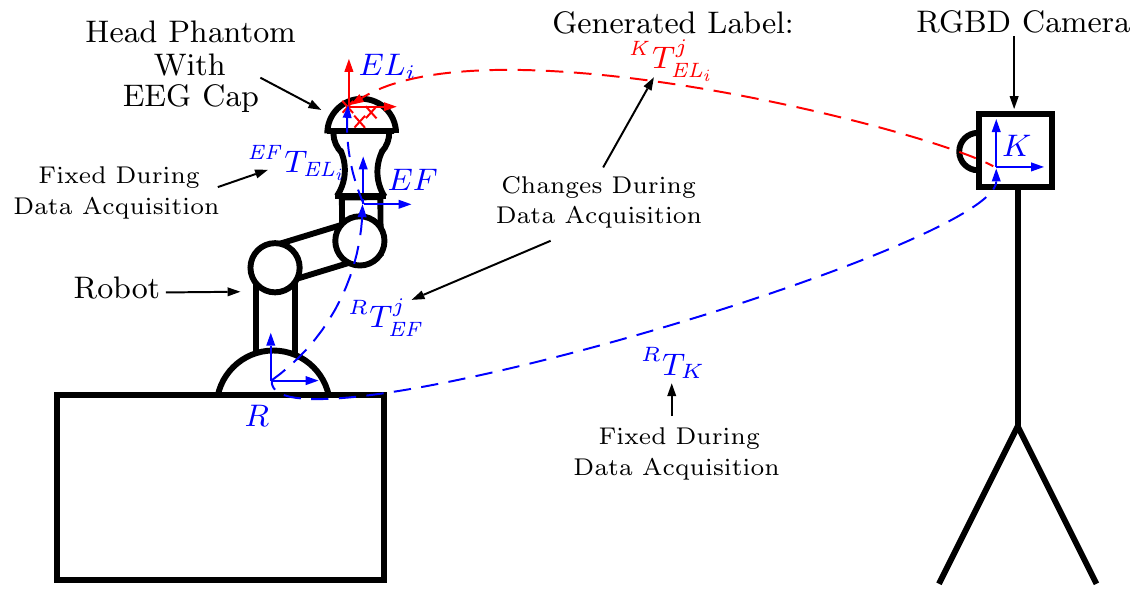}
	\caption{\textit {The experimental setup we use for data acquisition and label generation. Red crosses represent electrodes on the EEG cap.}}
	\label{fig:setup}
\end{figure}

Our experimental setup for data acquisition is shown in Figure~\ref{fig:setup}. First, we perform a hand-eye calibration between the robot (UR3, Universal Robots) and the camera (Kinect V2, Microsoft) using a checkerboard mounted to the robot. Camera poses of the checkerboard are obtained with OpenCV \cite{opencv_library} and the calibration transformations are obtained with QR24 \cite{ernst2012non} using $50$ robot and camera poses. Then, the head phantom wearing the electrode cap (waveguard touch, eemagine) is mounted to the robot. We now move the robot into different endeffector positions and orientations while continuously acquiring RGB and depth images while also logging the endeffector poses.

\subsection{Automatic Data Annotation}

After acquiring a set of images, we map the RGB images to the depth sensor's coordinate frame using calibrations provided by the manufacturer. The RGB images now have the same resolution as the depth images ($524\times 424$). Next, we annotate all electrodes in a single image. As the RGB image was transformed to the depth sensor's coordinate frame, we can now obtain the 3D coordinates from the 2D depth image and its corresponding point cloud using the 2D pixel locations from the RGB image. Using this 3D position and assuming identity orientation we obtain the poses $^KT_{\mathit{EL_i}}$ of all $N$ electrodes $i \in N$. Using the hand-eye calibration between robot and camera $^RT_K$ and the current endeffector pose $^RT_{\mathit{EF}}$ we obtain the electrode pose with respect to the robot endeffector:

\begin{equation}
    ^{\mathit{EF}}T_{\mathit{EL_i}} = (^RT_{\mathit{EF}})^{-1} \, ^{R}T_{K} \, ^{K}T_{\mathit{EL_i}}
\end{equation}

Next, we can automatically obtain the electrode poses for all other robot poses $j$ in the dataset:

\begin{equation}
    ^{\mathit{K}}T_{\mathit{EL_{i}}}^{j} = (^{R}T_{K})^{-1} \, ^{R}T_{\mathit{EF}}^{j} \, ^{\mathit{EF}}T_{\mathit{EL_{i}}}
\end{equation}

The position of the pose $^{\mathit{K}}T_{\mathit{EL_{i}}}^j$ is now used as a 3D label for each electrode position for each RGBD image in the dataset. Besides the 3D labels, we also consider image-level (pixel) labels by projecting the 3D points back into the RGB images. Here, we perform nearest-neighbor matching, i.e., we assign the electrodes' transformed 3D location to the closest point in the point cloud. Then, we project this point back on the RGB image.
While the 3D labels are ultimately used for EEG, the pixel labels can be useful for purely image-based algorithms. The automatically generated labels will likely be affected by calibration errors. Thus, we also compare to a more accurate ground-truth by manually labeling a small set of images. To ensure consistent labels, the annotator selects the center pixel of each electrode.  

\subsection{Deep Learning Models}

We employ two state-of-the-art CNNs, Densenet121 \cite{huang2017densely} and SE-Resnext50 \cite{hu2018squeeze}. The input to the network is an RGB image, the depth image or a full RGBD image. The images are cropped to the relevant region around the robot workspace based on the extent of the ground-truth annotations. Including a margin, this results in a network input size of $270 \times 254$ pixels. All models are pretrained on ImageNet to overcome relatively small dataset sizes. Using an EEG cap with $N=8$ electrodes, the model output is of size $N\times 3$ for 3D point prediction and $N\times 2$ for 2D pixel location prediction. As we solve a regression problem, the loss is the mean squared error, minimized using the Adam algorithm with an initial learning rate of $l_r = \num{e-5}$ and a batch size of $b=10$. We train for $300$ epochs and halve the learning after $50$ epochs each. For implementation we use PyTorch \cite{paszke2017automatic}. Training, evaluation and inference time measurement is performed on an NVIDIA GTX1080 TI.

In terms of evaluation metrics we follow \cite{borchani2015survey} and use the mean absolute error (MAE) as an absolute metric, either in pixels or $\si{\milli\metre}$ for 2D and 3D positions, respectively. To compare 2D and 3D labels, we consider the relative MAE (rMAE) which is the absolute error divided by the targets' standard deviation. The metric does not have a unit as it is relative. Last, we consider the average correlation coefficient (aCC) between predictions and targets as a relative metrics. Values close to $1$ indicate that a regression task was generally learned well.

Note that our fixed-size CNN output always forces the CNN to make a prediction for all electrode locations, even when they are not visible. Also, our automatic labeling strategy can provide annotations for learning even if some electrode locations are not visible as they are still transformed to their corresponding 3D location. Thus, our model is given the capability to obtain robustness towards partial electrode occlusion.

\section{\hspace{14pt}Results}

\begin{table}
	\centering
	\begin{tabular}{l l l l l l l l l l}
	 & & \multicolumn{3}{c}{Generated Labels} & & \multicolumn{3}{c}{Manual Labels} & \\
	 & & MAE & rMAE $(\num{e-3})$ & aCC & & MAE & rMAE $(\num{e-3})$ & aCC \\ \hline 
	\parbox[t]{2mm}{\multirow{7}{*}{\rotatebox[origin=c]{90}{2D Labels}}} & Gen. Labels & - & - & - & & $1.17 \pm 0.99$ & $38 \pm 2$ & $0.999$  \\ 
    & DN RGB & $\bm{0.71 \pm 0.59}$ & $25 \pm 2$ & $0.999$ & & $\bm{1.30 \pm 1.5}$ & $43 \pm 5$ & $0.998$ \\	
	& DN RGBD & $0.73 \pm 0.61$ & $25 \pm 2$ & $0.999$ & & $1.31 \pm 1.5$ & $43 \pm 5$ & $0.998$ \\
	& DN D & $1.50 \pm 1.3$ & $53 \pm 5$ & $0.998$ & & $1.91 \pm 1.8$ & $64 \pm 6$ & $0.997$ \\
	& SR RGB & $0.72 \pm 0.64$ & $25 \pm 2$ & $0.999$ & & $1.31 \pm 1.5$ & $44 \pm 5$ & $0.998$ \\
	& SR RGBD & $0.72 \pm 0.65$ & $25 \pm 2$ & $0.999$ & & $1.32 \pm 1.5$ & $44 \pm 5$ & $0.998$ \\
	& SR D & $1.56 \pm 1.3$ & $55 \pm 5$& $0.998$ & & $2.00 \pm 1.8$ & $67 \pm 6$ & $0.997$ \\\hline 
	\parbox[t]{2mm}{\multirow{7}{*}{\rotatebox[origin=c]{90}{3D Labels}}} & Gen. Labels & - & - & - & & $4.76 \pm 5.1$ & $51 \pm 2$ & $0.985$  \\  
    & DN RGB & $3.54 \pm 2.9$ & $62 \pm 5$ & $0.997$ & & $6.24 \pm 6.4$ & $131 \pm 19$ & $0.981$ \\	
	& DN RGBD & $3.82 \pm 3.3$ & $65 \pm 6$ & $0.996$ & & $6.35 \pm 6.3$ & $132 \pm 19$ & $0.982$ \\
	& DN D & $5.43 \pm 4.5$ & $92 \pm 8$ & $0.994$ & & $7.24 \pm 6.6$ & $146 \pm 20$ & $0.979$ \\
	& SR RGB & $3.19 \pm 2.7$ & $56 \pm 5$ & $0.997$ & & $6.00 \pm 6.1$ & $127 \pm 19$ & $0.983$ \\
	& SR RGBD & $\bm{3.13 \pm 2.7}$ & $56 \pm 5$ & $0.997$ & & $\bm{5.69 \pm 6.1}$ & $123 \pm 19$ & $0.983$ \\
	& SR D & $5.43 \pm 4.8$ & $92 \pm 8$ & $0.994$ & & $7.18 \pm 6.7$ & $143 \pm 19$ & $0.980$ \\ \hline \\	
	\end{tabular}
	\caption{Results for all experiments with Densenet121 (DN) and SE-Resnext50 (SR). We consider the MAE, the rMAE, and aCC. For 3D poses, the MAE is given in $\si{\milli\metre}$, for 2D poses, the MAE is given in pixels. rMAE and aCC are relative metrics without unit. Generated labels refer to our automatic annotations with our robotic setup. Manual labels refer to manual annotation by a human.}
	\label{tab:results}
\end{table}

To evaluate our setup we generate a set of $3000$ images for training and validation and we manually annotate $150$ images for testing. The positions cover a range of $\SI{500 x  500 x 300}{\milli\metre}$ which corresponds to $200 \times 200$ pixels in 2D images.

Using $40$ pairs of poses for calibration and $10$ pairs for evaluation, the hand-eye calibration between the robot and the camera results in a position error of $\SI[separate-uncertainty = true]{3.91 \pm 0.007}{\milli\metre}$ and a rotation error of $\SI[separate-uncertainty = true]{0.78 \pm 0.004}{\degree}$.

Quantitative results are shown in Table~\ref{tab:results}. We provide errors with respect to both the generated labels and the manually annotated labels. In general, our automatically generated labels are close to the real labels. Comparing Densenet121 and SE-Resnext50, both models perform similar with Densenet121 showing the best performance on 2D labels and SE-Resnext50 showing the best performance on 3D labels. With respect to color channels, using RGB and RGBD images performs similar. Both our pixel labels and the real-world 3D coordinates are learned well by the CNNs with aCCs close to $1$. For pixel labels, the relative metrics indicate a higher performance than for real-world 3D coordinates. Inference times are $\SI{34}{\milli\second}$ for Densenet121 and $\SI{58}{\milli\second}$ for SE-Resnext50. Training times are $\SI{3}{\hour}$ and $\SI{5}{\hour}$ for Densenet121 and SE-Resnext50, respectively.

\begin{figure}
	\centering
	\includegraphics [width=0.42\linewidth]{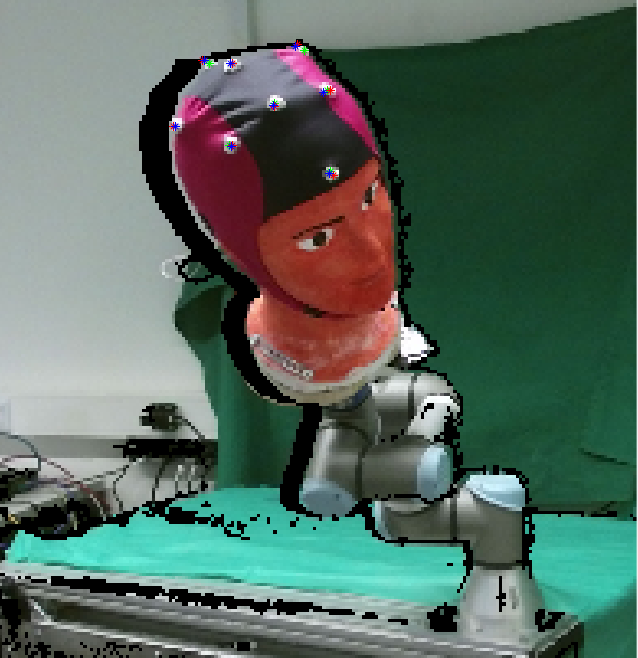}
	\includegraphics [width=0.52\linewidth]{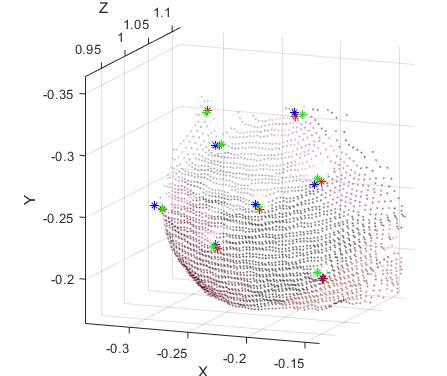}
	\caption{\textit {Left, a cropped RGB image with 2D electrode locations is shown. The black pixels are artifacts caused by the transformation of the RGB image to the depth sensor coordinate frame. Right, a point cloud with a zoom on the EEG cap with 3D electrode locations is shown. Green marks indicate the manually annotated electrode locations. Red marks indicate the automatically generated labels. Blue marks indicate the model predictions. Note that the 2D and 3D locations were predicted by different models.}}
	\label{fig:qual_results}
\end{figure}

Qualitative results are shown in Figure~\ref{fig:qual_results}. We show an RGB image and a point cloud of the head with the EEG cap and the manually annotated electrode locations, the generated labels and the predicted locations. Qualitatively, the predicted electrode locations are close to the manually annotated labels. Also, note that our approach is able to provide a reasonable prediction although one of the electrodes at the back of the head is only partially visible.

\section{\hspace{14pt}Discussion}

In this paper we address deep learning-based electrode detection using 2D camera images. This approach is particular promising as CNNs can provide fast predictions and they can be adjusted to versatile environments without requiring manual feature handcrafting. However, their main drawback is the large amount of annotated data that is usually required. We address this issue with a robotic setup for automatic data and label acquisition. We evaluate the approach by using different types of input images, labels and CNN architectures.

In general, our automatic label generation framework works well although the setup is affected by calibration errors between the robot and the camera. The generated labels closely match the more accurate manual ground-truth with an MAE of $\SI[separate-uncertainty = true]{4.76 \pm 5.1}{\milli\metre}$ and $1.17 \pm 0.99$ pixels while our labels cover a range of approximately $\SI{500 x 500 x 300}{\milli\metre}$ and $200 \times 200$ pixels. Also the point cloud plots in Figure~\ref{fig:qual_results} demonstrate that the actual electrode locations are well matched. Overall, the deep learning models approximate the ground-truth well, although there is a large variation in the target positions. Notably, there is a performance difference between using the generated and the manual labels for evaluation. This reflects the calibration errors in the setup which mainly cause the difference between generated and manual labels. Thus, the error between the generated labels and the manual labels can be seen as an upper bound for model performance.

At the same time, predictions are fast with a range of $\SI{34}{\milli\second}$ to $\SI{58}{\milli\second}$ which indicates real-time capability. Other photogrammetric methods typically require seconds up to minutes for detection \cite{clausner2017photogrammetry}.

Using either pixel or real-world 3D coordinates works well while predicting 2D labels appears to be easier with an average aCC of $0.998$ compared to $0.981$. Intuitively, deriving 3D coordinates from a 2D image is more difficult and thus the results match expectations. In terms of application, the 3D coordinates are more relevant as the overall goal is to obtain the electrode locations with respect to a 3D head coordinate frame. Adding head coordinates for deriving a head coordinate frame is straight forward with our approach and could be addressed in future work.

In terms of CNN models, the performance with respect to the actual labels is very similar as both models achieve aCCs close to $1$. SE-Resnext50 performs slightly better with respect to the 3D labels while Densenet121 shows the best performance for 2D pixel labels. Notably, the task of predicting 3D real-world coordinates appears to be more difficult as the rMAE and aCC are generally lower for this task. Considering that SE-Resnext50 has more parameters than Densenet121, the additional capacity might be beneficial for solving the more difficult problem. However, the slight increase in performance is bought with a substantial increase in inference time which needs to be carefully traded off for application. 

For the different types of input modalities the performance with respect to the manual labels is very similar. Adding the depth channel to the RGB images does not appear to be beneficial in our setup. This appears to indicate that depth information is not helpful, however, our current setup utilizes a particular head shape and EEG cap model. When generalizing to different head sizes, shapes or EEG caps, depth information will be more important as it is difficult to differentiate between a smaller head, close to the camera and a larger head, further away. Our automatic acquisition and labeling approach is well suited for covering more diverse scenarios, e.g., with different head models which can be addressed in future work. 

Overall, we demonstrate that predicting electrode locations works well and enables studying deep learning methods for EEG electrode detection. The setup allows for considering variations such as head shape and size or different EEG caps. However, a clear drawback of our method is the fact that it is limited to the use of head phantoms for automatic label generation. Thus, future work needs to incorporate our approach in real-world settings. This could be facilitated by applying a pre-segmentation of the electrode cap which would make the approach independent of the underlying head. Another approach would be to employ transfer learning techniques \cite{oquab2014learning} or few-shot learning \cite{sung2018learning} where a CNN that is pretrained with our setup is adapted to the real-world scenarios with a few new images.

\section{\hspace{14pt}Conclusion}

We study deep learning-based EEG electrode detection from camera images. To facilitate the approach, we propose a robotic setup for automatic data and label generation. This allows for quick generation of arbitrarily-sized datasets including ground-truth annotations. We demonstrate that CNNs are able to detect the electrodes using either RGB or depth images. Furthermore, our automatically generated labels closely match a more accurate, manual ground-truth annotation. Thus, our setup allows for developing and studying deep learning-based electrode detection approaches. Future work could study more extended scenarios, e.g., with different head shapes and sizes.

\section{\hspace{14pt}Acknowledgements}

This work was partially funded by AiF research grant number ZF4026302CR7.

\bibliographystyle{spmpsci} 
\small      
\bibliography{egbib}

\end{document}